\documentclass{article} 
\usepackage[T1]{fontenc}
\usepackage{iclr2026_conference,times}

\usepackage{amsmath,amsfonts,bm}

\def\eqref#1{equation~\ref{#1}}

\def\1{\bm{1}}

\DeclareMathAlphabet{\mathsfit}{\encodingdefault}{\sfdefault}{m}{sl}
\SetMathAlphabet{\mathsfit}{bold}{\encodingdefault}{\sfdefault}{bx}{n}

\usepackage{amssymb} 
\usepackage{hyperref}
\usepackage{url}
\usepackage{times}
\usepackage{latexsym}
\usepackage{amsmath}
\usepackage{amsfonts}
\usepackage{bbm}
\usepackage{tabularx}
\usepackage{booktabs}            
\usepackage[table]{xcolor}   
\usepackage{xcolor}
\usepackage{multirow}            
\usepackage{array}  
\usepackage{fancyhdr}
\usepackage{graphicx}
\usepackage{wrapfig} 
\definecolor{lightpink}{RGB}{255,220,230}
\usepackage{soul}
\sethlcolor{lightpink}
\usepackage[most]{tcolorbox}

\definecolor{sectbg}{RGB}{246,247,255}
\definecolor{goodgreen}{RGB}{0,128,0}
\definecolor{badred}{RGB}{180,0,0}
\definecolor{neutralgray}{RGB}{100,100,100}
\newcommand{\tred}[1]{\textcolor{goodgreen}{\scriptsize$\downarrow$\!#1\%}}
\newcommand{\aup}[1]{\textcolor{goodgreen}{\scriptsize$\uparrow$\!#1}}
\newcommand{\adn}[1]{\textcolor{badred}{\scriptsize$\downarrow$\!#1}}
\newcommand{\anc}[1]{\textcolor{neutralgray}{\scriptsize$\approx$}}

\title{Reasoning Under Constraint: How Batch Prompting Suppresses Overthinking in Reasoning Models}

\newcommand{\afflogo}[1]{\raisebox{-0.2ex}{\includegraphics[height=.8em]{#1}}}

\author{
\makebox[\textwidth][c]{%
\hspace{-0.05\textwidth}
\begin{tabular}{@{}c@{}}
Saurabh Srivastava\afflogo{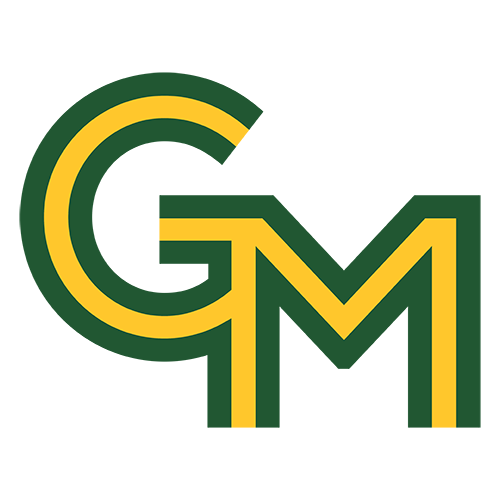} \hspace{0.8em}
Janit Bidhan\afflogo{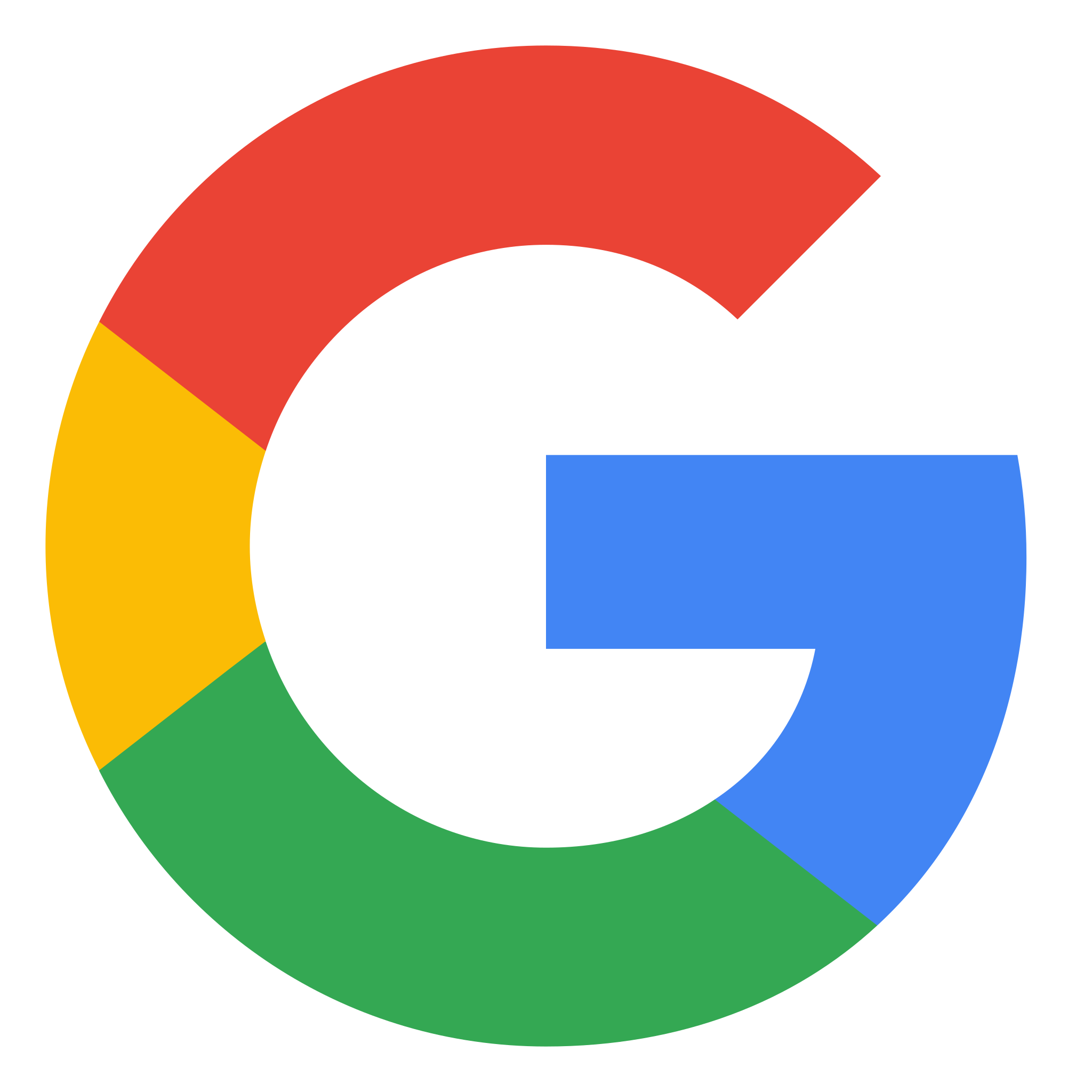} \hspace{0.8em}
Hao Yan\afflogo{gm.png} \hspace{0.8em}
Abhishek Dey\afflogo{gm.png} \hspace{0.8em}
Tanu Kansal\afflogo{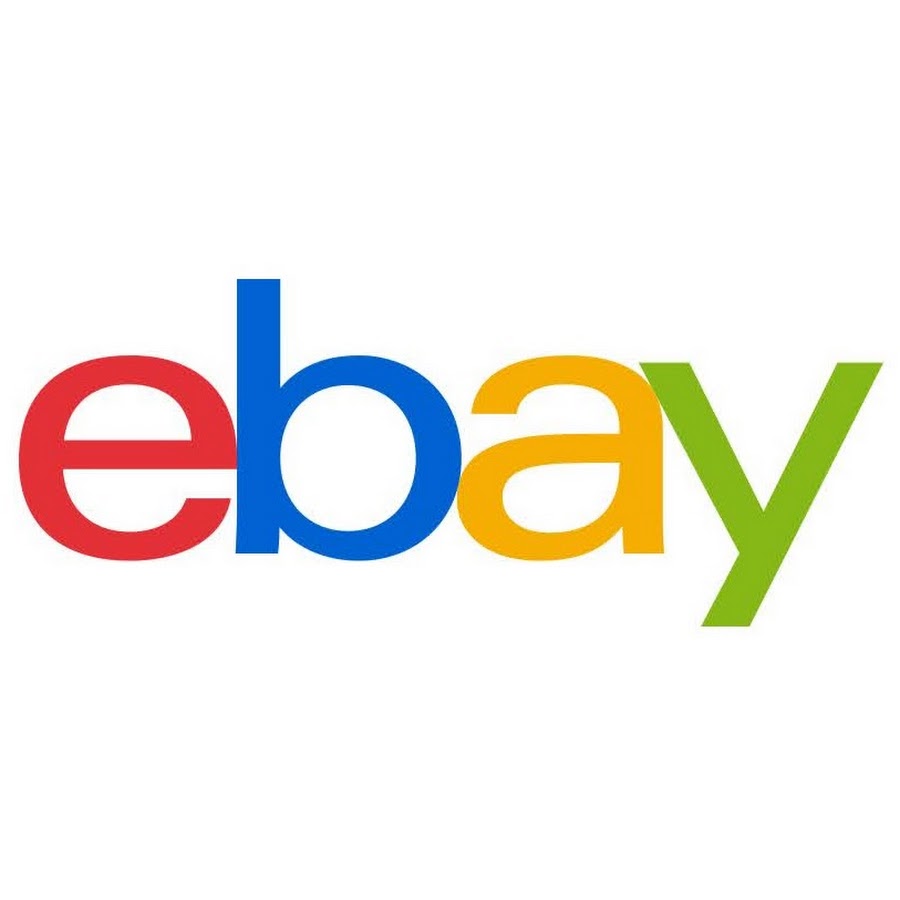} \\
Paras Kath\afflogo{gm.png} \hspace{0.8em}
Sina Mansouri\afflogo{gm.png} \hspace{0.8em}
Mohit Marvania\afflogo{gm.png} \hspace{0.8em}
Vamsi Shankar Simhadri\afflogo{gm.png} \hspace{0.8em}
Gaurav Singh\afflogo{gm.png} \\
[2pt]
{\normalfont
\afflogo{gm.png}\;George Mason University \hspace{1.2em}
\afflogo{google.png}\;Google \hspace{1.2em}
\afflogo{ebay.jpg}\;eBay}
\end{tabular}
}%
}

\iclrfinalcopy 
\begin{document}

\maketitle

\begin{abstract}
Large Reasoning Models (LRMs) achieve strong performance through explicit chain-of-thought reasoning but suffer from \textit{overthinking}: generating excessive reasoning tokens even for trivial queries. {Beyond inflating cost, overthinking can be self-defeating: models enter recursive self-doubt loops that exhaust token budgets without producing an answer, causing API timeouts that directly hurt accuracy.} We present an empirical study showing that \textbf{batch prompting}, originally introduced for throughput optimization, effectively suppresses overthinking at inference time. Across 13 diverse benchmarks with DeepSeek-R1 and OpenAI-o1, batch prompting {reduces reasoning tokens by 76\% (2{,}950$\mapsto$710), on average, while preserving or improving accuracy}. Through behavioral analysis, we find that batching induces three beneficial effects: (1) it reduces per-query reasoning effort when multiple queries share a context; (2) it enables pattern induction, where models generalize from earlier examples to solve later ones; and (3) it suppresses hedging behavior (e.g., ``\texttt{wait,}'' ``\texttt{let me double-check}'') that signals metacognitive loops. We also show that explicit prompt constraints (``\texttt{Use no more than 100 tokens in thinking.}'') fail to reduce overthinking; models either ignore them or sacrifice accuracy. These findings reframe batch prompting as more than a cost optimization: it is a practical inference-time technique that improves efficiency and reliability without model modification\footnote{Code, prompts, and all model outputs will be made publicly available upon acceptance.}.
\end{abstract}

\section{Introduction}
\begin{wrapfigure}{r}{0.5\textwidth} %
\vspace{-10mm}
  \centering
  \includegraphics[width=0.5\textwidth]{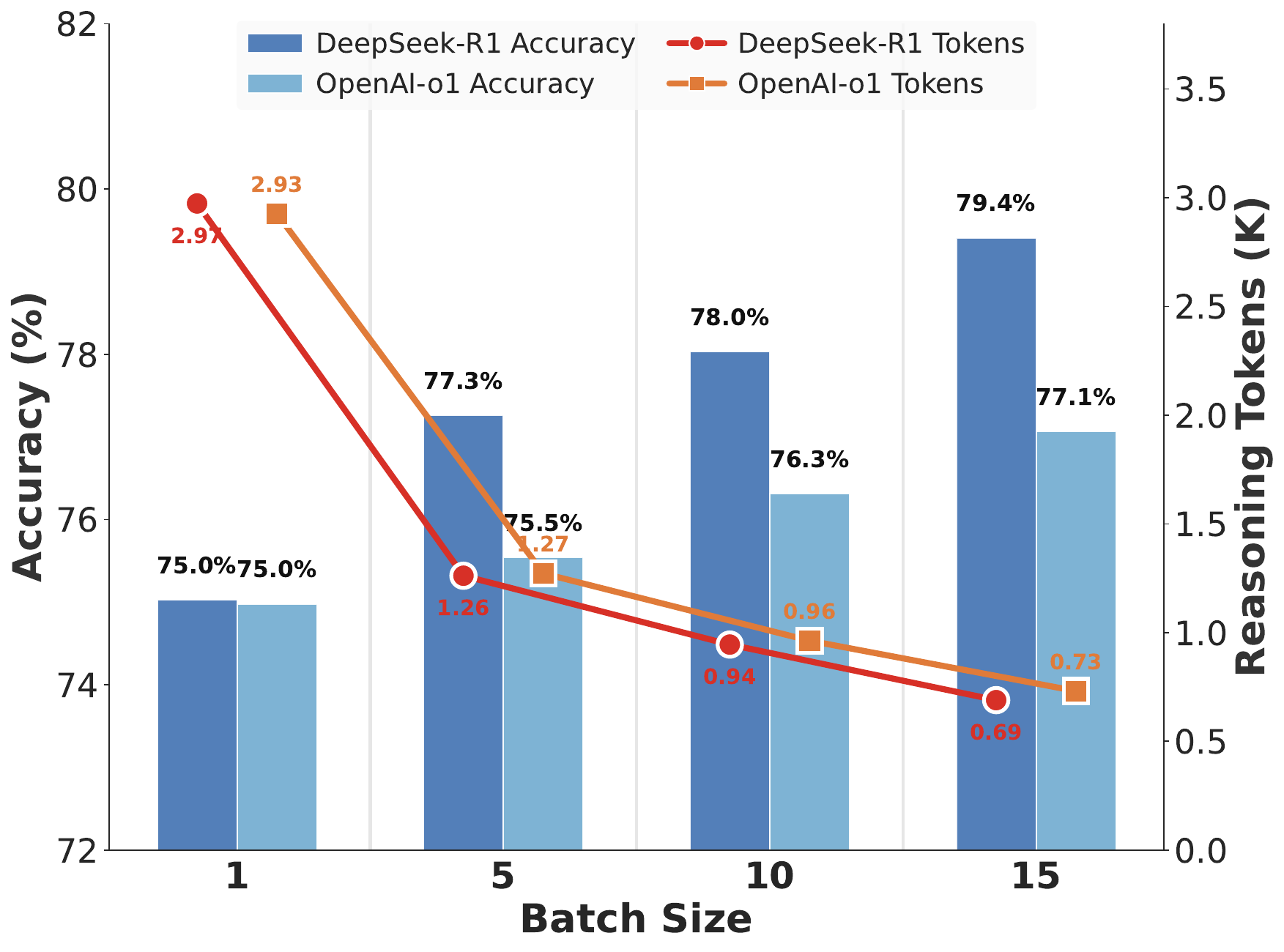} %
  \caption{
  Results for DeepSeek-R1 and OpenAI-o1 (averaged over 13 datasets) demonstrate that \textit{batching dramatically reduces reasoning tokens without hurting accuracy}. Increasing batch size improves accuracy while achieving a 76\% reduction in reasoning tokens.
}
    \label{fig:poc}
  \vspace{-1mm}
\end{wrapfigure}


Large Reasoning Models (LRMs) have achieved state-of-the-art performance in mathematics, code generation, and logical planning by producing explicit chains-of-thought (CoT) that expose intermediate reasoning steps~\citep{zhong2024evaluation, guo2025deepseek}. However, this deliberate reasoning comes at a cost: models frequently \textbf{overthink}, generating thousands of reasoning tokens even for trivial queries~\citep{cuadron2025dangeroverthinkingexaminingreasoningaction, huang2025mitigatingoverthinkinglargereasoning, chen2025think23overthinkingo1like, srivastava2025revisiting}. For API-based deployments where reasoning tokens are billed at premium rates, this behavior renders LRMs economically impractical at scale. {Worse, overthinking is not merely wasteful; it is accuracy-degrading. Models can enter recursive self-doubt loops that exhaust their token budget without producing an answer, causing API timeouts that directly reduce task accuracy.}

Prior approaches to curbing overthinking require model-internal access: self-training methods teach early stopping~\citep{wei2026evolutionthoughttrackingllm}, while activation-steering~\citep{huang2025mitigatingoverthinkinglargereasoning} suppresses token generation by modifying hidden states. These strategies are inapplicable to closed-weight models such as OpenAI-o1. This leaves a fundamental gap: \textit{can we reduce overthinking entirely at inference time, without touching the model?}


We answer affirmatively with a surprising finding: {batch prompting}~\citep{cheng-etal-2023-batch}, originally introduced for throughput optimization, acts as an effective \textit{implicit regularizer} that suppresses overthinking. When multiple queries are presented together, the model distributes its \textit{reasoning effort} across all questions rather than exhaustively analyzing each one, analogous to how humans abbreviate reasoning when juggling multiple tasks under time pressure. We hypothesize that this arises because the shared context creates a \textit{soft behavioral constraint}: the model allocates fewer reasoning tokens per query as the number of concurrent queries increases.


Figure~\ref{fig:poc} quantifies this effect. {Increasing the batch size from 1 to 15 reduces the reasoning tokens by \textbf{76\%} (2,950~$\mapsto$~710 tokens), on average, while preserving or improving accuracy.} This $\sim$4.5$\times$ reduction in reasoning tokens is achieved with \textit{zero} model modification, only a change in how queries are presented. The effect is consistent across both DeepSeek-R1~\citep{guo2025deepseek} and OpenAI-o1~\citep{zhong2024evaluation}, suggesting a consistent pattern in how reasoning models allocate computation under shared context.

Beyond efficiency, we observe that batching induces several beneficial behavioral changes. Models exhibit \textit{pattern induction}: when solving similar queries together, they generalize patterns from earlier examples to harder ones in the same batch. Batching also \textit{suppresses hedging}, tokens like ``\texttt{wait,}'' ``\texttt{let me double-check}'' that signal metacognitive loops, drop dramatically (from 21 occurrences $\mapsto$ 1 in our analysis). These effects suggest that batching acts as more than a cost optimization: it behaviorally regularizes the reasoning process.

{We also find that, in certain cases, single-query overthinking escalates to the point of complete failure: the model exhausts its token budget re-deriving and second-guessing an already-correct answer, producing no output before the API times out. This partly explains why single-query accuracy can fall below that of batched inference; batch prompting prevents such runaway generation by distributing reasoning effort across queries (Section~\ref{sec:analysis}).}

We validate these findings through a comprehensive study across 13 diverse benchmarks spanning arithmetic reasoning, question answering, structured extraction, and scientific tasks. Our contributions are as follows:
\begin{enumerate}
    \item \textbf{Batch prompting suppresses overthinking:} We show that batch prompting, beyond its known efficiency benefits, {effectively suppresses overthinking in LRMs at inference time, reducing reasoning tokens by 76\% while preserving or improving accuracy, without model modification.}

    \item \textbf{Comprehensive empirical validation:} We validate across 13 benchmarks, 2 frontier models (DeepSeek-R1, OpenAI-o1), and 4 batch sizes, establishing batch prompting as a robust, model-agnostic technique for efficient LRM deployment.

    \item \textbf{Behavioral analysis:} We provide a detailed analysis of \textit{how} batching regularizes reasoning: through pattern induction, hedging suppression, and adaptive token allocation that preserves accuracy on difficult instances.

    \item \textbf{Mechanistic hypotheses:} We propose candidate explanations for why batching suppresses overthinking, including shared-context pressure on reasoning depth, sequential anchoring from earlier responses, and implicit difficulty calibration, and discuss how our empirical findings relate to each.
\end{enumerate}

\section{Related Works}
\textbf{Overthinking in LRMs.} Large Reasoning Models frequently generate unnecessarily long reasoning traces, a phenomenon termed ``overthinking'' by \citet{chen2025think23overthinkingo1like}. They show diminishing returns from extended chains-of-thought, with models producing 10$\times$ more tokens than necessary on simple tasks. \citet{cuadron2025dangeroverthinkingexaminingreasoningaction} demonstrates that overthinking actively harms performance through error accumulation, while \citet{huang2025mitigatingoverthinkinglargereasoning} proposes activation-steering to suppress excessive generation. However, these model-internal approaches are inapplicable to closed-weight APIs like OpenAI-o1. Most closely related to our work, \citet{yan2025drqadynamicreasoningquota} observes that batch processing induces implicit resource competition in reasoning models, leading to more efficient reasoning allocation. They leverage this insight to generate preference data from batch outputs and train models via reinforcement learning to replicate this adaptive behavior during single-question inference. While conceptually similar in recognizing the benefits of batching, their approach requires model fine-tuning and access to model weights. Our work demonstrates that the benefits of batch-induced reasoning compression can be obtained \textit{directly at inference time} without any training, making it immediately applicable to closed API settings.

\textbf{Batch Prompting.} \citet{cheng-etal-2023-batch} introduced batch prompting as a throughput optimization, showing that grouping queries reduces per-query cost through prompt overhead amortization. Their focus was on standard Large Language Models (LLMs) and cost efficiency. We discover an orthogonal benefit: for \textit{reasoning} models specifically, batching serves as an implicit regularizer that suppresses overthinking---an effect not observed or characterized in prior work.

\textbf{Prompt-Based Efficiency.} Zero-shot prompting techniques like self-consistency \citep{wang2022self} and least-to-most prompting \citep{zhou2022least} aim to improve accuracy, but often incur added generation cost. More recent work reduces CoT length via output constraints or refinement. Our work operates in a black-box prompt-only setting, demonstrating that batching provides a natural, effective mechanism for reasoning brevity without explicit token constraints.

\section{Methodology and Formal Setup}
\begin{wrapfigure}{r}{0.5\textwidth} %
\vspace{-12mm}
  \centering
  \includegraphics[width=.8\linewidth]{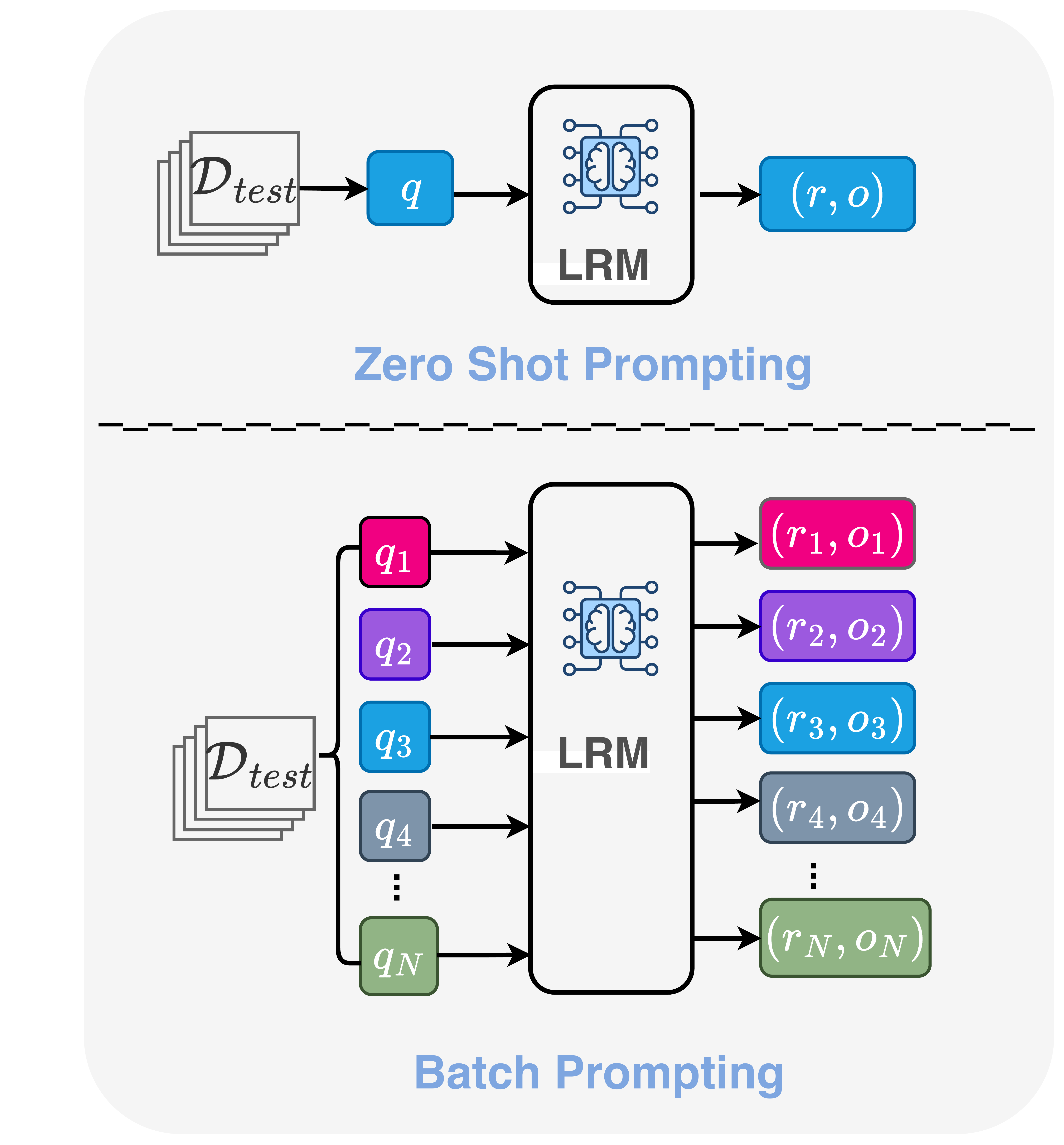} %
  \caption{
  {Batch prompting mechanism. Multiple queries ($q=[q_1,\ldots,q_N]$) are processed jointly, forcing the model to distribute attention and computation across all items. This implicit constraint suppresses overthinking without explicit instruction. The final responses are produced with a combination of reasoning and output tokens $(r_i, o_i)$}.
}
    \label{fig:overview}
  \vspace{-1mm}
\end{wrapfigure}
We consider the problem of minimizing inference-time cost for multi-step reasoning tasks without compromising model accuracy. Let $\mathcal{M}$ be a reasoning-oriented language model (e.g., DeepSeek-R1 or OpenAI-o1) that consumes prompt tokens and emits a \textit{response} comprising reasoning (thinking) tokens and output (answer) tokens. For a single query $q$, the total token cost consists of fixed prompt overhead $T_{\text{fix}}$, query tokens $T_q$, reasoning tokens $T_r$, and output tokens $T_o$. Thus, the total token usage is:
\[
C(q) = T_{\text{fix}} + T_q + T_r + T_o~.
\]
This cost is incurred independently for each query in traditional inference, leading to inefficiencies, especially in high-throughput applications. Crucially, for reasoning-oriented models, $T_r \gg T_o$ in practice, i.e., reasoning traces often consume 5--10$\times$ more tokens than the final answer, making $T_r$ the primary target for cost reduction.

\subsection{Batch Prompting for Efficiency}

To reduce this cost, we apply \textbf{batch prompting}~\citep{cheng-etal-2023-batch}, which groups $b$ queries into a single prompt (Figure~\ref{fig:overview}):
\[
P = \texttt{[Instruction]} \;||\; q_1 \;||\; q_2 \;||\; \cdots \;||\; q_b~,
\]
where $||$ denotes concatenation. The model processes $P$ in a single forward pass, producing responses $\{(r_1, o_1), \ldots, (r_b, o_b)\}$ where each response comprises reasoning tokens $r_i$ and output tokens $o_i$. The total cost for $b$ queries is then:
\[
C_{\text{batch}} = T_{\text{fix}} + \sum_{i=1}^b (T_{q_i} + T_{r_i} + T_{o_i})~.
\]
The effective per-query cost becomes:
\[
\frac{C_{\text{batch}}}{b} = \frac{T_{\text{fix}}}{b} + \frac{1}{b} \sum_{i=1}^b (T_{q_i} + T_{r_i} + T_{o_i})~,
\]
showing that prompt overhead is amortized as $1/b$, yielding substantial gains for large $b$. In practice, batching leads to {$\sim$76\%} reduction in token usage on average with minimal loss in accuracy (see Figure~\ref{fig:poc}).

\subsection{Why Does Batching Reduce Reasoning? Candidate Hypotheses}
\label{sec:why_batching}
The empirical observation that batching reduces reasoning tokens while preserving accuracy calls for explanation. We propose three candidate hypotheses that are consistent with our findings. We emphasize that these are \textit{conjectures} informed by our experiments, not proven mechanisms; disentangling their individual contributions remains an open question.

\paragraph{Hypothesis 1: Shared-Context Pressure.}
We conjecture that when $b$ queries share a single context window, the model faces implicit pressure to produce shorter per-query reasoning traces. In single-query inference, the model can devote its full generation budget to a single query, but with batching it must produce $b$ responses within the same generation, which may discourage extensive deliberation on any individual item. The precise mechanism behind this effect remains unclear and could involve learned generation-length priors, context-window utilization patterns, or other factors. Nevertheless, the empirical trend is consistent: per-query reasoning tokens decrease monotonically with batch size across both models (Figure~\ref{fig:poc}).

\paragraph{Hypothesis 2: Sequential Anchoring.}
When generating responses for batched queries, the model processes them sequentially within a single generation. Let $\hat{y}_i$ denote the full response to query $q_i$. The conditional generation follows:
\[
P(\hat{y}_i \mid q_1, \ldots, q_b, \hat{y}_1, \ldots, \hat{y}_{i-1})~.
\]
One possible explanation for the observed brevity is that if $\hat{y}_1$ adopts a concise reasoning style, subsequent responses $\hat{y}_2, \ldots, \hat{y}_b$ are biased toward similar brevity through in-context pattern matching. This is consistent with our observation that batching enables pattern induction (Section~5), though we have not isolated this effect from the shared-context hypothesis above.

\paragraph{Hypothesis 3: Implicit Difficulty Calibration.}
We hypothesize that the presence of multiple queries provides implicit calibration of task difficulty. When a model sees query $q_i$ alongside other queries $\{q_j\}_{j \neq i}$, the batch context may signal that exhaustive reasoning is not required for any single item. This is analogous to how humans adjust effort when faced with multiple tasks under time pressure. However, we note that our experiments use semantically homogeneous batches, which limits our ability to test this hypothesis against heterogeneous mixtures.

\medskip
\noindent
Regardless of the underlying mechanism, the empirical effect is robust. Let $R(q)$ denote the reasoning trace length. We observe:
\[
\mathbb{E}[|R(q)|_{\text{batch}}] \ll \mathbb{E}[|R(q)|_{\text{single}}]~,
\]
with the reduction factor increasing with batch size (Figure~\ref{fig:poc}). Importantly, this compression appears \textit{adaptive}: models allocate more tokens to harder queries within a batch while abbreviating easier ones, preserving accuracy despite reduced total computation.

\section{Experiments}
\textbf{Datasets and Metrics.}
{We evaluate across 13 diverse benchmarks spanning arithmetic (GSM8k~\citep{mirzadeh2024gsm}, Math500~\citep{hendrycks2021measuring}, Game of 24~\citep{yao2023tree}), factual and epistemic QA (GPQA~\citep{rein2024gpqa}, SciTab~\citep{lu2023scitab}, Epistemic~\citep{ghazal2013bigbench}, StrategyQA~\citep{geva2021did}), structured tasks (Event Extraction~\citep{doddington2004automatic}, Object Count~\citep{ghazal2013bigbench}, Last Letter Concatenation~\citep{ghazal2013bigbench}), visual or perceptual reasoning (Penguins~\citep{ghazal2013bigbench}), and challenging benchmarks such as IIT-JEE~\citep{arora-etal-2023-llms} and BBEH~\citep{kazemi2025big}.} We use 100 samples from each dataset, consistent with standard LLM evaluation practice~\citep{zhong2024evaluation}. Accuracy is measured via exact match or task-specific metrics, while efficiency is tracked using model-reported reasoning and output token counts. We conduct all experiments with DeepSeek-R1 (deepseek-reasoner, accessed via API) and OpenAI-o1 (o1-2024-12-17), using zero-shot templated prompting (see Appendix~\ref{prompt_template} for full prompt template). For both models, we use default API parameters. Batches are constructed by randomly sampling $b$ queries from each dataset and grouping them into a single prompt; all queries within a batch come from the same dataset (homogeneous batching). We evaluate batch sizes $b \in \{1, 5, 10, 15\}$. Token counts (reasoning and output) are obtained from the API response metadata. If a model exhausts its token budget without producing a parseable answer, we score that instance as incorrect. We provide additional details in Appendix~\ref{app:details}.

\noindent
\paragraph{Statistical Methodology.}
For token count and accuracy comparisons, we report means with 95\% confidence intervals (CIs) computed as $\bar{x} \pm 1.96 \cdot \frac{s}{\sqrt{n}}$. 
Confidence intervals are calculated over pooled evaluation instances ($n = 1300$ per model per batch size; 13 datasets $\times$ 100 samples each), since batching is applied uniformly and all metrics are defined at the instance level. 
Pooling across instances provides a stable estimate of the population-level effect of batching while preserving statistical validity. 
We observe that reasoning token reductions are consistent in direction across all 13 datasets and both models, indicating robustness beyond any single-dataset comparison. 
Full confidence interval tables and additional statistical details are provided in Appendix~\ref{sec:additional_res}.


\definecolor{sectbg}{RGB}{246,247,255}
\definecolor{avgcol}{RGB}{233,240,255}

\newcommand{\secttitle}[1]{%
  \rowcolor{sectbg}\multicolumn{18}{c}{\bfseries #1}\\[-0.3ex]
  \arrayrulecolor{sectbg!40!black}\specialrule{0.5pt}{0pt}{0pt}\arrayrulecolor{black}
}

\begin{table*}[t!]
\centering
\setlength{\tabcolsep}{3pt}
\renewcommand{\arraystretch}{1.08}

\resizebox{\linewidth}{!}{%
\begin{tabular}{l|cccc|ccccccc|ccccc|>{\columncolor{avgcol}}c}
\toprule

\rowcolor{sectbg}\multicolumn{18}{c}{\textbf{All results reported with \textit{OpenAI-o1}}}\\[-0.6ex]
\midrule

\multirow{2}{*}{\textbf{Batch Size}} &
\multicolumn{4}{c|}{\textbf{Info Extraction}} &
\multicolumn{7}{c|}{\textbf{Reasoning / QA}} &
\multicolumn{5}{c|}{\textbf{Structured / Scientific}} &
\multirow{2}{*}{\textbf{Average}} \\
\cmidrule(lr){2-5}\cmidrule(lr){6-12}\cmidrule(lr){13-17}
& TI & TC & AI & AC
& Penguins & GSM8K & StratQA & IIT & GPQA & Last & Epis.
& ObjCnt & Math & BBEH & SciTab & Game24 & \\
\midrule

\secttitle{Accuracies (\%)}
1  & 58.40 & 54.70 & 31.80 & 29.80 & 99.00 & 97.00 & 89.00 & 87.00 & 79.00 & 95.00 & 84.00 & 99.00 & 88.00 & 42.00 & 70.00 & 96.00 & {74.98} \\
5  & 55.80 & 50.60 & 25.20 & 23.00 & 100.00 & 96.00 & 91.00 & 93.00 & 83.00 & 97.00 & 91.00 & 100.00 & 91.00 & 45.00 & 70.00 & 97.00 & {75.54} \\
10 & 59.70 & 57.10 & 29.40 & 25.70 & 99.00 & 97.00 & 92.00 & 91.00 & 81.00 & 97.00 & 90.00 & 99.00 & 92.00 & 46.00 & 69.00 & 96.00 & {76.31} \\
15 & 64.40 & 59.10 & 33.90 & 30.70 & 98.00 & 96.00 & 91.00 & 91.00 & 82.00 & 96.00 & 91.00 & 99.00 & 91.00 & 43.00 & 71.00 & 96.00 & {\textbf{77.07}} \\
\addlinespace[2pt]
\midrule

\secttitle{Reasoning Tokens}
1  & 1570.80 & 1694.56 & 1830.50 & 1876.89 & 1393.14 & 1792.57 & 1166.58 & 3701.48 & 7559.94 & 1788.95 & 1229.70 & 2680.54 & 9815.81 & {4531.56} & {1825.05} & 2361.40 & {2926.22} \\
5  & 459.45 & 486.67 & 585.45 & 583.36 & 340.66 & 575.46 & 288.92 & 3178.64 & 3223.71 & 499.13 & 577.85 & 1206.04 & 4668.14 & {2527.85} & {512.15} & 637.53 & {1271.94} \\
10 & 430.58 & 440.67 & 559.56 & 560.56 & 260.36 & 562.37 & 279.35 & 2929.45 & 2333.25 & 271.89 & 418.56 & 934.24 & 2959.47 & {1829.60} & {278.98} & 352.47 & {962.58} \\
15 & 427.78 & 432.67 & 430.76 & 427.89 & 229.57 & 505.24 & 268.45 & 2735.75 & 1313.87 & 170.59 & 384.45 & 734.35 & 2184.48 & {1030.26} & {175.04} & 246.44 & {\textbf{731.10}} \\
\addlinespace[2pt]
\midrule

\secttitle{Output Tokens}
1  & 430.80 & 449.54 & 451.67 & 458.56 & 242.07 & 264.18 & 239.73 & 742.87 & 8135.57 & 201.18 & 241.28 & 395.58 & 615.22 & {504.48} & {232.29} & 407.17 & {875.76} \\
5  & 205.56 & 208.56 & 210.67 & 211.43 & 144.07 & 179.31 & 149.09 & 337.40 & 275.18 & 134.34 & 172.60 & 179.58 & 211.33 & {173.29} & {161.14} & 202.01 & {197.22} \\
10 & 198.25 & 196.49 & 192.22 & 189.45 & 127.52 & 172.57 & 142.23 & 318.46 & 205.74 & 108.94 & 157.58 & 173.65 & 205.74 & {168.71} & {139.92} & 157.58 & {178.44} \\
15 & 175.27 & 171.21 & 165.45 & 161.34 & 127.58 & 169.35 & 135.73 & 294.36 & 119.74 & 74.61 & 129.43 & 133.21 & 194.25 & {159.28} & {107.12} & 131.21 & {\textbf{153.07}} \\
\bottomrule
\end{tabular}%
}
\caption{Accuracies, reasoning tokens, and output tokens across 13 benchmarks grouped by task type. TI, TC, AI and AC represent Trigger Identification, Trigger Classification, Argument Identification and Argument Classification tasks on ACE'05.}
\label{tab:all_metrics_batched_o1}
\vspace{0.3cm} 
\end{table*}

\begin{table*}[t!]
\centering
\setlength{\tabcolsep}{3pt}
\renewcommand{\arraystretch}{1.08}

\resizebox{\linewidth}{!}{%
\begin{tabular}{l|cccc|ccccccc|ccccc|>{\columncolor{avgcol}}c}
\toprule

\rowcolor{sectbg}\multicolumn{18}{c}{\textbf{All results reported with \textit{DeepSeek-R1}}}\\[-0.6ex]
\midrule

\multirow{2}{*}{\textbf{Batch Size}} &
\multicolumn{4}{c|}{\textbf{Info Extraction}} &
\multicolumn{7}{c|}{\textbf{Reasoning / QA}} &
\multicolumn{5}{c|}{\textbf{Structured / Scientific}} &
\multirow{2}{*}{\textbf{Average}} \\
\cmidrule(lr){2-5}\cmidrule(lr){6-12}\cmidrule(lr){13-17}
& TI & TC & AI & AC
& Penguins & GSM8K & StratQA & IIT & GPQA & Last & Epis.
& ObjCnt & Math & BBEH & SciTab & Game24 & \\
\midrule

\secttitle{Accuracies (\%)}
1  & 59.17 & 55.14 & 32.66 & 30.50 & 99.00 & 98.00 & 90.00 & 88.00 & 80.00 & 96.00 & 85.00 & 97.00 & 42.00 & 80.00 & 71.00 & 97.00 & {75.03} \\
5  & 56.63 & 51.23 & 25.96 & 23.35 & 100.00 & 97.00 & 92.00 & 93.00 & 84.00 & 98.00 & 91.00 & 100.00 & 45.00 & 84.00 & 98.00 & 97.00 & {77.26} \\
10 & 59.83 & 57.58 & 29.63 & 26.37 & 98.00 & 96.00 & 92.00 & 92.00 & 81.00 & 97.00 & 96.00 & 99.00 & 44.00 & 86.00 & 98.00 & 96.00 & {78.03} \\
15 & 65.11 & 59.88 & 34.36 & 31.27 & 98.00 & 96.00 & 92.00 & 91.00 & 91.00 & 100.00 & 96.00 & 97.00 & 44.00 & 83.00 & 97.00 & 95.00 & {\textbf{79.41}} \\
\addlinespace[2pt]
\midrule

\secttitle{Reasoning Tokens}
1  & 1470.80 & 1607.56 & 1817.50 & 1845.89 & 1310.14 & 1777.57 & 1079.58 & 3666.48 & 7489.94 & 1767.95 & 1148.70 & 2580.54 & 9766.81 & {5759.42} & {2184.45} & 2301.40 & {2973.42} \\
5  & 412.45 & 440.67 & 537.45 & 499.36 & 301.66 & 550.46 & 248.92 & 3166.64 & 3148.71 & 426.13 & 553.85 & 1115.04 & 4588.14 & {3105.60} & {502.36} & 585.53 & {1261.44} \\
10 & 335.58 & 417.67 & 504.56 & 506.56 & 205.36 & 562.37 & 254.35 & 2885.45 & 2266.25 & 206.89 & 401.56 & 893.24 & 2887.47 & {2235.22} & {243.90} & 310.47 & {944.81} \\
15 & 387.78 & 335.67 & 387.76 & 335.89 & 165.57 & 432.24 & 217.45 & 2701.75 & 1233.87 & 137.59 & 334.45 & 657.35 & 2151.48 & {1216.98} & {162.20} & 171.44 & {\textbf{689.34}} \\
\addlinespace[2pt]
\midrule

\secttitle{Output Tokens}
1  & 413.80 & 431.54 & 401.67 & 366.56 & 221.07 & 213.18 & 165.73 & 714.87 & 8073.57 & 142.18 & 154.28 & 384.58 & 532.22 & {452.39} & {163.05} & 381.17 & {825.74} \\
5  & 151.56 & 133.56 & 150.67 & 106.07 & 112.31 & 127.09 & 274.40 & 213.36 & 258.18 & 56.34 & 159.60 & 153.58 & 144.33 & {122.68} & {118.77} & 155.01 & {152.34} \\
10 & 171.25 & 156.49 & 122.22 & 166.45 & 59.64 & 153.57 & 117.23 & 255.46 & 120.46 & 8.94 & 102.29 & 77.94 & 168.74 & {143.43} & {61.18} & 95.58 & {123.80} \\
15 & 139.27 & 159.21 & 142.45 & 64.34 & 79.58 & 116.35 & 83.73 & 213.36 & 65.74 & 3.61 & 34.43 & 109.45 & 159.25 & {135.36} & {20.92} & 77.21 & {\textbf{100.27}} \\
\bottomrule
\end{tabular}%
}
\caption{Accuracies, reasoning tokens, and output tokens across 13 benchmarks grouped by task type.  TI, TC, AI and AC represent Trigger Identification, Trigger Classification, Argument Identification and Argument Classification tasks on ACE'05.}
\label{tab:all_metrics_batched}
\end{table*}

\subsection{Main Results}
\noindent
\textbf{Batching Improves Accuracy Across Tasks.}
{We evaluate batch prompting across 13 diverse benchmarks spanning question answering, arithmetic reasoning, structured extraction, and perception-style tasks (Tables~\ref{tab:all_metrics_batched_o1}-\ref{tab:all_metrics_batched}). Compared to single-query prompting (batch size = 1), larger batches (up to 15) generally improve, though some tasks show non-monotonic behavior. For instance, on OpenAI-o1, ACE'05 trigger identification accuracy increases from 58.4 to 64.4, while on tasks prone to overthinking~\citep{srivastava-etal-2024-instances, rein2024gpqa}, batch prompting yields notable boosts (e.g., GPQA: 79 $\mapsto$ 83; Last Letter: 95 $\mapsto$ 97 on o1). Averaged across both models, accuracy rises from 75.0\% to 78.2\% at batch size 15. We hypothesize this gain stems from both token efficiency and the inductive signal offered by seeing similar queries together, reinforcing the right reasoning patterns.}

\noindent
\textbf{Batching Reduces Reasoning Cost with No Output Penalty.}
Beyond accuracy, batching substantially reduces reasoning tokens. Averaged across both models, reasoning cost drops from 2,950 to just 710 tokens (4.5$\times$ reduction), with a corresponding decrease in output tokens (851 $\mapsto$ 127).  Even in reasoning-heavy tasks like GSM8K or SciTab, batching reduces internal deliberation while preserving performance, consistent with the hypothesis that batching discourages excessive internal deliberation. This effect is especially prominent in tasks where models tend to hedge or overexplore (e.g., Game of 24, Epistemic QA). Together, these results demonstrate that batch prompting is a practical and general-purpose technique for improving LRM efficiency without model or system changes.

\definecolor{chedge}{HTML}{FFD4D4}     
\definecolor{crecheck}{HTML}{FFE8C0}   
\definecolor{crepeat}{HTML}{C8DEFF}    
\definecolor{ctangent}{HTML}{DDD0FF}   
\definecolor{panelAbg}{HTML}{EEF3FF}   
\definecolor{panelBbg}{HTML}{FFF0F0}   

\newcommand{\sst}{\rule[-1pt]{0pt}{8pt}}

\begin{table*}[t!]
\centering
\small
\setlength{\tabcolsep}{5pt}
\setlength{\fboxsep}{1.5pt}            
\renewcommand{\arraystretch}{1.3}

\resizebox{\textwidth}{!}{%
\begin{tabular}{@{} >{\centering\arraybackslash}m{1cm} p{18cm} @{}}
\toprule

\multicolumn{2}{@{}c@{}}{\cellcolor{panelAbg}\rule{0pt}{14pt}%
\textsc{\textbf{Part A:}\; Overthinking Patterns --- Standard vs.\ Batch Prompting}} \\
\midrule

\multirow{9}{*}{\rotatebox[origin=c]{90}{\parbox{5.5cm}{\centering
      \textbf{Standard Prompting}\\\textbf{(BS\,=\,1)}}}}

& $\blacktriangleright$ \textbf{Q1:} Which is larger, 0.9 or 0.11? \\[4pt]

& \quad $\triangleright$ \textit{Reasoning:}\; We need to compare 0.9 and 0.11.
  Obviously 0.9 is larger because \textbf{(\ldots)}
  So $0.9 > 0.11$.
  \colorbox{chedge}{\sst But maybe the question is a trick?}
  \colorbox{chedge}{\sst Sometimes people might} \textbf{(\ldots)}
  \colorbox{chedge}{\sst But no, it is decimal.}
  So the answer is 0.9.
  \textbf{(\ldots)}. \\[3pt]

& \quad $\Rightarrow$ \textit{Output:}\;
  0.9 is larger than 0.11.
  When comparing decimals, 0.9 is equivalent to 0.90,
  which is greater than 0.11. \\[8pt]
\cmidrule(l{2em}){2-2}

& $\blacktriangleright$ \textbf{Q2:} How many Rs in ``Strawberrrry''? \\[4pt]

& \quad $\triangleright$ \textit{Reasoning:}\; We need to count the number of `R's
  \textbf{(\ldots)}.
  Let's spell it: S-t-r-a-w-b-e-r-r-r-r-y \textbf{(\ldots)}.
  \colorbox{crecheck}{\sst Actually careful:}
  The word is ``Strawberrrry''.
  \colorbox{chedge}{\sst Typically} \textbf{(\ldots)}
  \colorbox{crecheck}{\sst Actually strawberry:} \textbf{(\ldots)}
  \colorbox{crepeat}{\sst However, lets double-check} \textbf{(\ldots)}
  \colorbox{chedge}{\sst Actually strawberry}
  So just count the \textbf{(\ldots)}. \\[3pt]

& \quad $\Rightarrow$ \textit{Output:}\;
  There are 5 Rs in ``Strawberrrry''. \\[8pt]
\cmidrule(l{2em}){2-2}

& $\blacktriangleright$ \textbf{Q3:} A fort has enough food for 100 soldiers
  for 120 days. 10~days later, 20 more soldiers join.
  How many more days will the food last? \\[4pt]

& \quad $\triangleright$ \textit{Reasoning:}\; We are given:
  ``A fort has enough food \textbf{(\ldots)}
  \colorbox{chedge}{\sst But the question} \textbf{(\ldots)}
  new soldiers join.
  \colorbox{crecheck}{\sst But sometimes} they want integer? \textbf{(\ldots)}
  \colorbox{chedge}{\sst Perhaps they mean} ``how many more''? \textbf{(\ldots)}
  \colorbox{chedge}{\sst But maybe they} want ``more days'' \textbf{(\ldots)}
  \colorbox{crepeat}{\sst Lets read carefully} ``10 days later \textbf{(\ldots)}
  \\[3pt]

& \quad $\Rightarrow$ \textit{Output:}\;
  After 10 days, the food consumed by 100 soldiers is
  $100 \times 10 = 1000$ soldier-days \textbf{(\ldots)} \\

\midrule

\multirow{3}{*}{\rotatebox[origin=c]{90}{\parbox{1.6cm}{\centering
      \textbf{Batch Prompting}\\\textbf{(BS\,=\,5)}}}}

& $\blacktriangleright$ \textbf{Prompt:}\;Solve these questions:
  Q1: How many Rs are in ``Strawberrrry''?\;\;
  Q2: Fort food problem (same as above).\;\;
  Q3: Which is larger, 0.9 or 0.11?\textcolor{gray!80}{\;\textit{...truncated...}} \\[4pt]

& \quad $\triangleright$ \textit{Reasoning:}\; We need to answer three questions
  \textbf{(\ldots)} ``Strawberrrry'' --- ``Straw'' + ``berrrry''
  \textbf{(\ldots)}
  \colorbox{crecheck}{\sst Actually} strawberry - typical spelling \textbf{(\ldots)}
  \colorbox{chedge}{\sst But careful}: ``Strawberry'' - it might \textbf{(\ldots)}
  So answer: 5.\;
  Q2: A fort has \textbf{(\ldots)}
  \colorbox{chedge}{\sst But perhaps they} \textbf{(\ldots)};
  Q3: 0.9 or 0.11? \textbf{(\ldots)}. \\[3pt]

& \quad $\Rightarrow$ \textit{Output:}\;
  Q1: 5 Rs.\;\;
  Q2: Total food $= 100 \times 120 = 12{,}000$ soldier-days \textbf{(\ldots)}.\;\;
  Q3: 0.9 is larger.\textcolor{gray!80}{\;\textit{...truncated...}} \\

\specialrule{0.8pt}{8pt}{2pt}
\specialrule{0.8pt}{2pt}{4pt}

\multicolumn{2}{@{}c@{}}{\cellcolor{panelBbg}\rule{0pt}{14pt}%
\textsc{\textbf{Part B:}\; Overthinking-Induced API Timeout}} \\
\midrule

\multirow{3}{*}{\rotatebox[origin=c]{90}{\parbox{4cm}{\centering
      \textbf{Standard Prompting}\\\textbf{(BS\,=\,1)}}}}

& $\blacktriangleright$ \textbf{Q:} Take the last letters of each word in
  ``Oswaldo Jos\'{e} Luis Sheldon Tara'' and concatenate them. \\[4pt]

& \begin{minipage}[t]{17.8cm}
$\triangleright$ \textit{Reasoning:}\;
\colorbox{crepeat}{\sst ``Oswaldo'': O-s-w-a-l-d-o. Last letter is ``o''.}
\colorbox{chedge}{\sst But is it lowercase or uppercase?}
\colorbox{ctangent}{\sst For concatenation, case might matter \textbf{(\ldots)}}
\colorbox{crepeat}{\sst ``Jos\'{e}'': J-o-s-\'{e}. Last character is `\'{e}'.}
\colorbox{ctangent}{\sst In string processing, `\'{e}' is a single Unicode character \textbf{(\ldots)}}
\colorbox{crecheck}{\sst Perhaps I should confirm the words.}
\colorbox{crepeat}{\sst For each word: Oswaldo$\to$`o',\, Jos\'{e}$\to$`\'{e}',\, Luis$\to$`s',\, Sheldon$\to$`n',\, Tara$\to$`a'.}
\par\smallskip
\textcolor{green!55!black}{%
\textit{--- correct answer ``o\'{e}sna'' first derived here ---}}
\par\smallskip
\colorbox{ctangent}{\sst In some contexts, people might consider the base letter \textbf{(\ldots)}}%
\colorbox{crecheck}{\sst Let me double-check.}%
\colorbox{crepeat}{\sst ``Oswaldo'': O,s,w,a,l,d,o $\to$ `o';}
\colorbox{crepeat}{\sst ``Jos\'{e}'': J,o,s,\'{e} $\to$ `\'{e}' (U+00E9).}
\colorbox{ctangent}{\sst In Python: words\,=\,[\ldots]}
\colorbox{chedge}{\sst I think for this context, it's fine.}
\textbf{(\ldots)}
\colorbox{chedge}{\sst but the word is given}
\textbf{(\ldots)} \\
\textcolor{gray!80}{\;\textit{continues for 5,000+ reasoning tokens}}
\end{minipage} \\[8pt]

& \hfill\fcolorbox{red!70!black}{red!8}{\;%
  \textbf{\textcolor{red!70!black}{%
  $\boldsymbol\Rightarrow$\; API Timeout --- No answer generated.}}\;} \\

\midrule
\multirow{3}{*}{\rotatebox[origin=c]{90}{\parbox{2.2cm}{\centering
      \textbf{Batch}\\\textbf{Prompting}\\\textbf{(BS\,=\,5)}}}}

& $\blacktriangleright$ \textbf{Q:} Solve these questions:
Q1: Take the last letters of each words in ``Oswaldo Jos\'{e} Luis Sheldon Tara'' and concatenate them.
Q2: Take the last letters of each words in ``Alec Arianna Corina Juancarlos'' and concatenate them. 
Q3: Take the last letters of each words in ``Leandro Maricela Genevieve Lesly'' and concatenate them. \textcolor{gray!80}{\;\textit{...truncated...}}\\[4pt]

& \begin{minipage}[t]{17.8cm}
\setlength{\parskip}{0pt}
\setlength{\spaceskip}{0pt}
$\triangleright$ \textit{Reasoning:}\;
We need to solve three questions. Each asks to take the last letters \textbf{(\ldots)}
Words Oswaldo, Jos\'{e}, Luis, Sheldon, Tara \textbf{(\ldots)}.
\colorbox{chedge}{But we need to consider if they want it as a single string.}
\textbf{(\ldots)}
Words: Alec, Arianna, Corina, and Juancarlos. 
\textbf{(\ldots)}
Words: Leandro, Maricela, Genevieve, Lesly \textbf{(\ldots)} 
\colorbox{chedge}{But we need to be careful:} The phrase in Q1 \textbf{(\ldots)}
Thus final answer Q1: o\'{e}sna;\; Q2: caas;\; Q3: oaey. \textcolor{gray!80}{\;\textit{...truncated...}}

\end{minipage} \\[3pt]

& \quad $\Rightarrow$ \textit{Output:}\;
Q1: o\'{e}sna;\;
Q2: caas;\;
Q3: oaey \textcolor{gray!80}{\;\textit{...truncated...}} \\[2pt]
\bottomrule
\end{tabular}
}%

\vspace{6pt}
{\footnotesize
\noindent\mbox{%
\textbf{Legend\,:}\;
\colorbox{chedge}{\sst\; Self-doubt / hedging\;}\;
\colorbox{crecheck}{\sst\; Redundant rechecking\;}\;
\colorbox{crepeat}{\sst\; Repetitive re-derivation\;}\;
\colorbox{ctangent}{\sst\; Irrelevant tangents\;}
}%
}

\caption{Qualitative comparison of reasoning behaviour in DeepSeek-R1
under standard vs.\ batch prompting.
\textbf{Part~A:} Single-query prompting triggers four distinct
overthinking patterns (see legend); batch prompting produces
concise reasoning for the \emph{same} queries with no model modification.
\textbf{Part~B:} In extreme cases, the model derives the correct answer
early but continues to spiral through
\colorbox{crepeat}{\scriptsize\sst repetition},
\colorbox{ctangent}{\scriptsize\sst tangents}, and
\colorbox{chedge}{\scriptsize\sst self-doubt}
until the API times out---directly explaining lower accuracy at BS\,=\,1. Batched examples are \textcolor{gray!80}{\;\textit{truncated}} showing only 3 examples to save space.}
\label{tab:token-usage-examples}
\end{table*}

\section{Analysis}
\label{sec:analysis}
\paragraph{Batch Prompting Regularizes Overthinking.}
{Despite their strength on multi-step tasks, LRMs frequently overthink even trivial problems, generating verbose and unstable reasoning traces.
Table~\ref{tab:token-usage-examples} (Part~A) illustrates this phenomenon across three simple queries: a decimal comparison, a letter-counting task, and an arithmetic word problem. In each case, DeepSeek-R1 produces the correct answer early in its reasoning trace but then spirals into self-doubt (``\texttt{But maybe the question is a trick?}''), redundant rechecking (``\texttt{Actually careful\ldots}''), and irrelevant tangents. We identify four distinct overthinking patterns: \textit{hedging, rechecking, repetitive re-derivation}, and \textit{irrelevant tangents}, that inflate token counts without improving answer quality.}

{\paragraph{Overthinking Hurts Accuracy, Not Just Efficiency.} 
In extreme cases, overthinking causes outright failure. Table~\ref{tab:token-usage-examples} (Part~B) shows a last-letter concatenation query where DeepSeek-R1 derives the correct answer within the first few hundred tokens, but then re-derives, second-guesses, and explores tangential concerns for over 5{,}000 tokens until the API times out with \textit{no answer generated}. This partly explains why single-query accuracy can fall below batched inference on certain tasks: overthinking can cause complete answer failures on instances the model is otherwise capable of solving. Batch prompting eliminates this failure mode: when the same query appears alongside similar examples (Table~\ref{tab:token-usage-examples} Part~B, bottom), the model answers all queries concisely without entering any reasoning spiral.}

\paragraph{Batching Reduces Overall Token Usage.}
\begin{wrapfigure}{r}{0.5\textwidth} %
\vspace{-5mm}
  \centering
  \includegraphics[width=\linewidth]{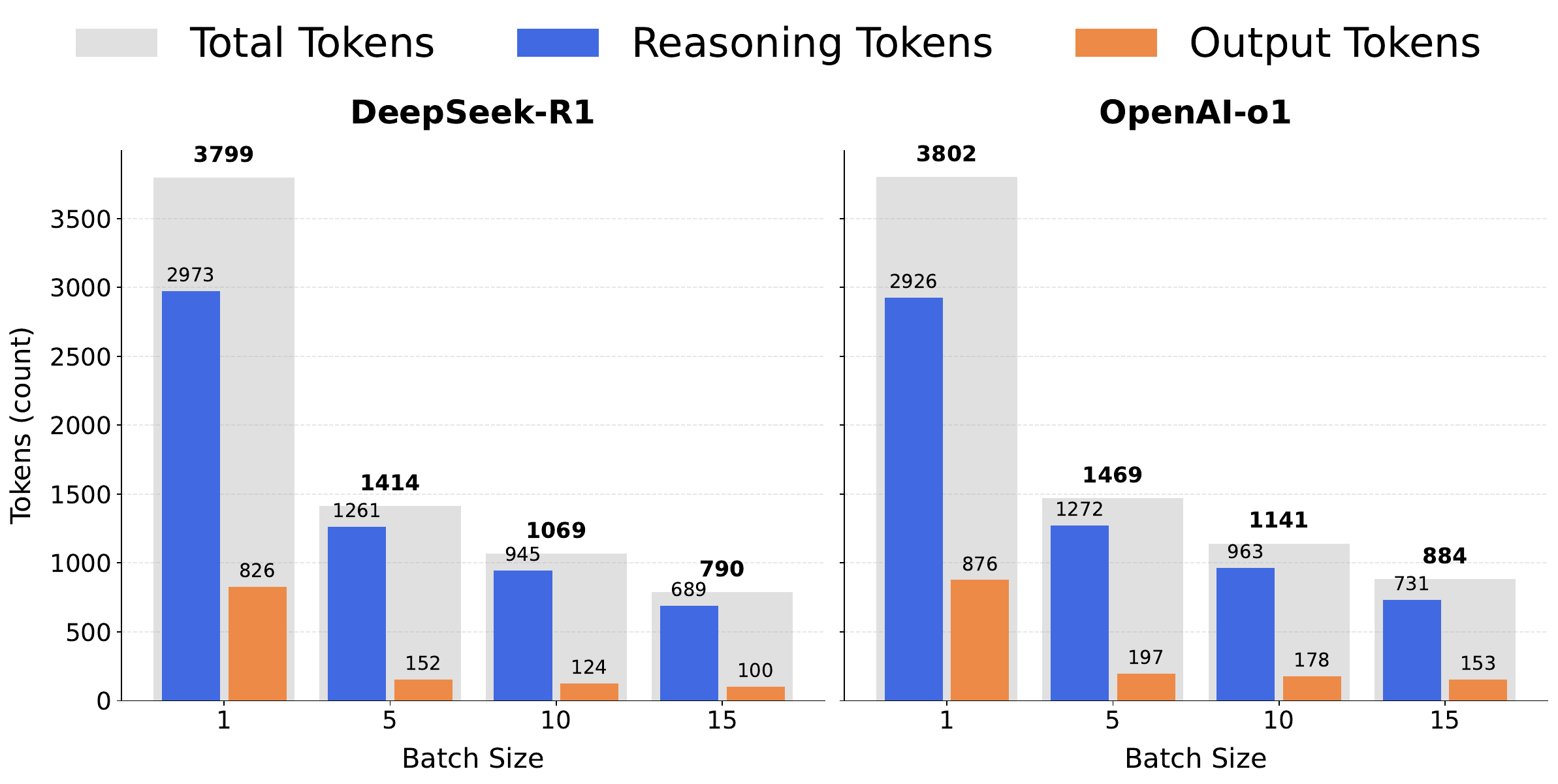} %
  \caption{Token breakdown for {DeepSeek-R1} and {OpenAI-o1}. Larger batch sizes reduce total tokens mainly by shortening reasoning traces, with minimal effect on output accuracies.}
    \label{fig:tok_usage}
  \vspace{-7mm}
\end{wrapfigure}
Figure~\ref{fig:tok_usage} decomposes per-query token consumption into reasoning and output components across batch sizes. Both models exhibit a monotonic decline in total tokens, but the savings are overwhelmingly driven by reasoning compression. For DeepSeek-R1, reasoning tokens drop from 2{,}973 (BS=1) to 689 (BS=15), a 76.8\% reduction, while output tokens decline from 826 to 100 (87.9\%). OpenAI-o1 follows the same trend: reasoning falls from 2{,}926 to 731 (75.0\%), output from 876 to 153 (82.5\%). In absolute terms, reasoning accounts for $\sim$76\% of the total token reduction at every batch size, confirming that batching primarily compresses the model's internal deliberation rather than truncating its answers. The resulting total cost drops from ${\sim}$3{,}800 tokens per query (BS=1) to ${\sim}$840 (BS=15), a 4.5$\times$ reduction, while accuracy improves (Tables~\ref{tab:all_metrics_batched_o1}--\ref{tab:all_metrics_batched}). While both components decrease substantially, reasoning tokens dominate the absolute savings because they constitute the majority of tokens at baseline (${\sim}$78\% of total). This is consistent with the view that batch prompting primarily compresses the model's internal deliberation, though we note that output tokens also decrease significantly (83--88\%).

\paragraph{Batching Encourages Pattern Induction.}
Beyond regularizing verbosity, batching structurally similar examples appears to facilitate inductive generalization. In isolation, models often struggle with formatting-sensitive tasks, e.g., when asked to extract and concatenate the last letters of words, DeepSeek-R1 produces the correct reasoning but outputs ``\texttt{nn yo}'' (with a spurious space), failing under exact match. However, when the same example is placed in a batch of similar queries, the model not only solves each instance correctly but also aligns to a more consistent output schema. This suggests that batched inference enables in-context pattern discovery, allowing the model to leverage structural similarity and prior completions to guide subsequent reasoning. In effect, later examples benefit from earlier ones, even without gradient updates, supporting the view that batches can serve as implicit demonstrations and reasoning anchors.

\paragraph{Batching Suppresses Hedging and Metacognitive Loops.}
\begin{wrapfigure}{r}{0.5\textwidth} %
\vspace{-5mm}
  \centering
  \includegraphics[width=\linewidth]{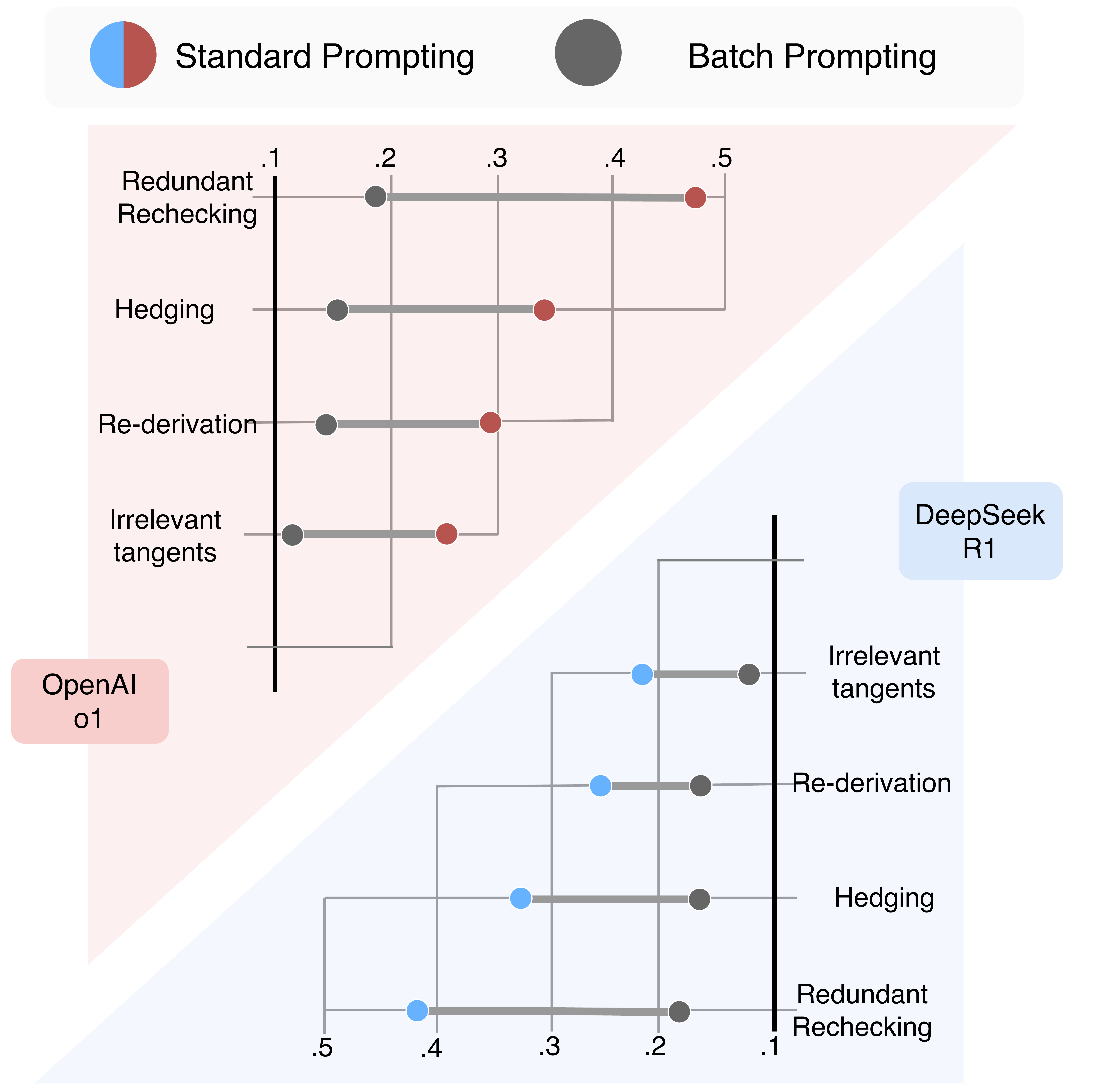} %
  \caption{Normalized frequency of four overthinking patterns under baseline (BS=1) vs.\ batch prompting (BS=5) for DeepSeek-R1 and OpenAI-o1. Batching uniformly suppresses all pattern types, with the largest reductions in rechecking and re-derivation.}
    \label{fig:hedging_count}
  \vspace{-5mm}
\end{wrapfigure}
One of the hallmarks of overthinking in LRMs is the emergence of verbose, recursive self-dialog, often marked by tokens like ``\texttt{wait,}'' ``\texttt{let me double-check,}'' or ``\texttt{hold on.}'' We quantify this by measuring the normalized frequency of four overthinking patterns across baseline (BS=1) and batched (BS=5) inference: \textit{hedging}, \textit{rechecking}, \textit{re-derivation}, and \textit{irrelevant tangents}. 
As shown in Figure~\ref{fig:hedging_count}, batching uniformly suppresses all four patterns across both models. DeepSeek-R1 exhibits the highest baseline frequencies (hedging: 0.42, rechecking: 0.55, re-derivation: 0.38, tangents: 0.31), which drop to 0.18, 0.22, 0.16, and 0.12, respectively, under batching, a 59\% reduction on average.

OpenAI-o1 shows a similar pattern: baseline frequencies of 0.33, 0.47, 0.29, and 0.24 fall to 0.15, 0.19, 0.13, and 0.11, representing a 56\% average reduction. Rechecking behavior is most prevalent at baseline and also most effectively suppressed, dropping by 60\% for DeepSeek-R1 and 60\% for OpenAI-o1. For instance, on the Game of 24 task, single-query inference with DeepSeek-R1 produced 21 instances of the token ``\texttt{wait}'' across five queries, reflecting repeated self-corrections; the same five questions in a batch yielded only one occurrence, focused solely on the most difficult query. This suggests that batching acts as a soft interrupt mechanism: by distributing attention across multiple queries, the model avoids getting \emph{stuck} in internal loops, implicitly curtailing metacognitive wandering.

\paragraph{Explicit Prompt Constraints Also Fail.}
A natural baseline is to ask whether simple prompt-level constraints can induce comparable reductions in overthinking. We evaluate two such interventions, explicit token limits (e.g., ``\texttt{use no more than 100 tokens for reasoning}'') and time limits (e.g., ``\texttt{think for at most 10 seconds}''), on \textsc{Game of 24} and \textsc{Last Letter}. As shown in Table~\ref{tab:prompt_comparison}, these approaches yield limited and inconsistent gains. Models frequently acknowledge the constraint in natural language but continue to generate long internal reasoning traces. For instance, on \textsc{Game of 24}, DeepSeek-R1 under an explicit token limit still produces 1{,}256 reasoning tokens on average, corresponding to only a 45\% reduction relative to the baseline (2{,}301 tokens), far below the 93\% reduction achieved by batch prompting. OpenAI-o1 is even less responsive: token constraints reduce reasoning tokens by only 17\% while substantially degrading accuracy (96\%\,$\rightarrow$\,45\%). 

These results highlight a key limitation of instructional constraints: they rely on the model to voluntarily comply, and can be ignored or superficially satisfied without materially reducing computation. In contrast, batch prompting creates a \emph{behavioral} constraint; the shared context appears to induce shorter reasoning traces without requiring explicit compliance, yielding large reductions in reasoning tokens while preserving, or even improving, task accuracy.

\section{Implications for Future Research}
Our findings that batch prompting compresses reasoning traces while preserving accuracy have implications beyond the evaluation setting studied in this paper. We outline several concrete directions where batched inference could yield practical gains.

\paragraph{Compact Chain-of-Thought Generation for Distillation.}
A major bottleneck in chain-of-thought (CoT) distillation~\citep{hsieh2023distilling} is that teacher models produce verbose reasoning traces, making the distillation data expensive to generate and harder for student models to learn from. Our results show that batch prompting produces shorter yet equally accurate reasoning traces (76\% fewer tokens at BS=15 with no accuracy loss). These compressed traces could serve as higher-quality supervision for training smaller models: the student receives the same correct reasoning signal in a fraction of the tokens, reducing both the cost of generating distillation data and the sequence length the student must learn to produce. Concretely, distilling from batched DeepSeek-R1 traces (averaging 689 reasoning tokens per query at BS=15) rather than single-query traces (2,973 tokens) would reduce the teacher-generation cost by ${\sim}$4.5$\times$ while providing more focused, less noisy supervision.

\begin{wraptable}{b}{0.55\textwidth}
\centering
\small
\setlength{\tabcolsep}{4pt}
\vspace{-10pt}
\begin{tabular}{@{}lccc@{}}
\toprule
\textbf{Dataset} & \textbf{Method} & \textbf{Accuracy} & \textbf{Avg. Tokens} \\
\midrule
\multicolumn{4}{c}{\centering\cellcolor{sectbg}\textbf{DeepSeek-R1}} \\
\midrule
\multirow{6}{*}{Game of 24}
  & Baseline (BS=1)         & 97.0\%                 & 2301 \\
  & + Token Limit           & 85.0\% \adn{12}         & 1256 \tred{45} \\
  & + Time Limit            & 96.0\% \adn{1}          & 1424 \tred{38} \\ \cmidrule{2-4}
  & \textbf{Batch (BS=5)}   & \textbf{97.0\%} \anc    & \textbf{586} \tred{75} \\
  & \textbf{Batch (BS=10)}  & \textbf{96.0\%} \adn{1} & \textbf{310} \tred{87} \\
  & \textbf{Batch (BS=15)}  & \textbf{95.0\%} \adn{2} & \textbf{171} \tred{93} \\
\midrule
\multirow{6}{*}{Last Letter}
  & Baseline (BS=1)         & 96.0\%                 & 1768 \\
  & + Token Limit           & 92.0\% \adn{4}          & 852  \tred{52} \\
  & + Time Limit            & 94.0\% \adn{2}          & 974  \tred{45} \\
  \cmidrule{2-4}
  & \textbf{Batch (BS=5)}   & \textbf{98.0\%} \aup{2} & \textbf{426} \tred{76} \\
  & \textbf{Batch (BS=10)}  & \textbf{97.0\%} \aup{1} & \textbf{207} \tred{88} \\
  & \textbf{Batch (BS=15)}  & \textbf{100.0\%} \aup{4}& \textbf{138} \tred{92} \\
\midrule

\multicolumn{4}{c}{\cellcolor{sectbg}\textbf{OpenAI-o1}} \\
\midrule

\multirow{6}{*}{Game of 24}
  & Baseline (BS=1)         & 96.0\%                 & 2361 \\
  & + Token Limit           & 45.0\% \adn{51}         & 1952 \tred{17} \\
  & + Time Limit            & 89.0\% \adn{7}          & 2125  \tred{10} \\\cmidrule{2-4}
  & \textbf{Batch (BS=5)}   & \textbf{97.0\%} \aup{1} & \textbf{638} \tred{73} \\
  & \textbf{Batch (BS=10)}  & \textbf{96.0\%} \anc    & \textbf{352} \tred{85} \\
  & \textbf{Batch (BS=15)}  & \textbf{96.0\%} \anc    & \textbf{246} \tred{90} \\
\midrule

\multirow{6}{*}{Last Letter}
  & Baseline (BS=1)         & 95.0\%                 & 1789 \\
  & + Token Limit           & 77.0\% \adn{18}         & 1607 \tred{10} \\
  & + Time Limit            & 92.0\% \adn{3}          & 1680  \tred{06}
  \\\cmidrule{2-4}
  & \textbf{Batch (BS=5)}   & \textbf{97.0\%} \aup{2} & \textbf{499} \tred{72} \\
  & \textbf{Batch (BS=10)}  & \textbf{97.0\%} \aup{2} & \textbf{272} \tred{85} \\
  & \textbf{Batch (BS=15)}  & \textbf{96.0\%} \aup{1} & \textbf{171} \tred{90} \\
\bottomrule
\end{tabular}
\caption{Comparison of prompt-based solutions for reducing overthinking. Explicit constraints modestly reduce tokens but \textit{can hurt accuracy} (red arrows), while batch prompting achieves large token reduction while \textit{preserving or improving accuracy} (green/gray arrows). Subscripts show change vs.\ baseline (BS=1).}
\label{tab:prompt_comparison}
\vspace{-15pt}
\end{wraptable}

\paragraph{Batched Prompt Optimization.}
In prompt optimization frameworks such as DSPy~\citep{khattab2023dspy} and OPRO~\citep{yang2024large}, an LLM evaluates candidate prompts by running inference on individual examples and using the outputs as a proxy gradient signal to update the prompt. This single-instance evaluation is both costly and noisy: each gradient step requires a separate API call, and the feedback from a single example may not generalize. Batch prompting offers a natural improvement: by evaluating a candidate prompt on a batch of $b$ examples in a single call, the optimizer receives $b$ data points of feedback simultaneously. This not only reduces the per-example API cost (as shown in our experiments), but may also yield more informative gradient signals, observing a prompt's behavior across multiple examples in one pass exposes systematic failure patterns that single-instance evaluation would miss. Recent work on prompt optimization with reasoning models~\citep{srivastava2025revisiting} has highlighted the cost barrier of iterative prompt refinement with LRMs; batched evaluation directly addresses this by reducing the number of API calls needed per optimization step by a factor of $b$.

\paragraph{Cost-Efficient Agentic Workflows.}
Multi-step agentic systems~\citep{yao2023react, shinn2023reflexion} invoke reasoning models repeatedly, for planning, tool selection, self-reflection, and error correction, making reasoning token costs a dominant expense. When an agent must process multiple independent sub-tasks (e.g., evaluating several candidate actions, verifying multiple facts, or generating plans for parallel sub-goals), these sub-tasks can be batched into a single prompt. Our finding that batching reduces reasoning tokens by $\sim$4.5$\times$ without accuracy loss suggests that batched sub-task processing could substantially reduce the cost of agentic pipelines. For instance, a ReAct-style agent~\citep{yao2023react} that must evaluate 5 candidate tool calls could batch these evaluations into a single prompt, reducing reasoning cost while obtaining all evaluations simultaneously.

\paragraph{Test-Time Compute Allocation.}
Recent work has shown that scaling test-time compute, allocating more inference-time reasoning, can be more effective than scaling model parameters~\citep{snell2024scaling}. However, current test-time scaling strategies treat each query independently, allocating compute without cross-query awareness. Our observation that batching induces adaptive token allocation (Section~\ref{sec:why_batching}), where models spend more tokens on harder queries and fewer on easier ones within a batch, suggests a complementary approach: batch prompting could serve as a lightweight mechanism for adaptive test-time compute allocation across a workload, without requiring explicit difficulty classifiers or routing logic. Combining batching with explicit compute-allocation strategies (e.g., adaptive batch sizes based on estimated query difficulty) is a promising direction for efficient deployment of reasoning models at scale.

\section{Conclusion}
We present an empirical study demonstrating that batch prompting effectively suppresses overthinking in Large Reasoning Models at inference time. Across 13 diverse benchmarks with DeepSeek-R1 and OpenAI-o1, batching reduces reasoning tokens by 76\% while maintaining or improving accuracy, a 4$\times$ efficiency gain with no model modification. Through behavioral analysis, we find that batching suppresses hedging and rechecking patterns, enables pattern induction from earlier examples, and adaptively allocates tokens to harder queries.

We also find that explicit prompt constraints (token/time limits) fail to reliably reduce reasoning tokens while often hurting accuracy, highlighting the advantage of batching's implicit approach. 
These findings reframe batch prompting as more than a throughput optimization: it is a practical, model-agnostic tool for efficient LRM deployment. Understanding the precise mechanisms behind batch-induced reasoning compression remains an open direction for future work.

\section{Limitations}
Our work has several limitations. First, we focus on batches of \textit{independent} queries; dependent queries (e.g., multi-turn dialogues) may benefit less from batching as models must maintain state across queries. Second, extreme batch sizes or highly heterogeneous query mixtures may degrade coherence; our experiments use semantically homogeneous batches. Third, while we use 100 samples per dataset (a standard for LLM evaluation), larger-scale validation would strengthen statistical claims; the consistency across 13 datasets partially addresses this. Finally, batching lacks explicit control over token budgets; future work should explore adaptive methods combining batching with dynamic reasoning control~\citep{srivastava-etal-2024-instances, srivastava-etal-2025-instruction}, as well as extensions to multi-modal settings~\citep{dey2025multimodal}.


\bibliography{iclr2026_conference}
\bibliographystyle{iclr2026_conference}

\appendix
\section{Prompt Template}
\label{prompt_template}
\begin{tcolorbox}[
  colback=gray!5,
  colframe=black!60,
  title={Prompt Template},
  fonttitle=\bfseries,
  arc=2mm,
  boxrule=0.6pt,
  left=6pt,right=6pt,top=6pt,bottom=6pt
]
You are a highly intelligent and capable AI assistant. Your job is to solve the user queries given below. 

Answer Format: \texttt{\{answer\_format\}}

You can reason about all your questions together in \texttt{<reasoning>...</reasoning>} block.

Then, for each question, your job is to solve each question and provide your response in the following format:
\texttt{<answers>}
{\footnotesize
\[
\begin{array}{l}
[\\
\quad \{\{"example 1": \{\{"reason": "Let's think step by step. ... ", "answer": "answer 1"\}\}\}\},\\
\quad \{\{"example 2": \{\{"reason": "Let's think step by step. ... ", "answer": "answer 2"\}\}\}\},\\
\quad \ldots\\
\quad \{\{"example n": \{\{"reason": "Let's think step by step. ... ", "answer": "answer n"\}\}\}\}\\
]\\
\end{array}
\]
}%
\texttt{</answers>}

For each example, start with "Let's think step by step. ... " in the reason field. For each answer, provide the final answer in the answer field and nothing else.

Here are the questions you need to answer:
\texttt{\{batch\_prompt\}}

\medskip
\textbf{---\textgreater\ }Remember, you have to solve \texttt{\{num\_questions\}} questions in total so plan in advance to answer cheaply and efficiently. \texttt{\{constrain\}}\textbf{\ \textless---}

\begin{tcolorbox}[
  colback=blue!5,
  colframe=blue!60!black,
  title={Constraints},
  fonttitle=\bfseries,
  arc=1.5mm,
  boxrule=0.5pt
]
{\texttt{time constrain: "Think for only 10 seconds."}}

\texttt{token constrain: "Use no more than 100 tokens in thinking."}
\end{tcolorbox}

\end{tcolorbox}

\section{Experimental Details}
\label{app:details}

\paragraph{Models and API Configuration.}
We use two reasoning-oriented LLMs: (1) \textbf{DeepSeek-R1} (model ID: \texttt{deepseek-reasoner}), accessed via the DeepSeek API, and (2) \textbf{OpenAI-o1} (model ID: \texttt{o1-2024-12-17}), accessed via the OpenAI API. Both models expose reasoning (thinking) tokens separately from output tokens in their API responses. Temperature and top-p are not user-configurable for these reasoning endpoints; we use default settings throughout.

\paragraph{Batch Construction.}
For each dataset, we randomly sample 100 instances and partition them into batches of size $b \in \{1, 5, 10, 15\}$. All queries within a batch are drawn from the same dataset (homogeneous batching). The order of queries within each batch is randomized. For batch size 1, we run standard single-query inference as the baseline.

\paragraph{Output Parsing and Scoring.}
Batched model outputs are parsed by extracting the structured JSON response for each query (delimited by the \texttt{<answers>} tags in the prompt template). If a query's response cannot be parsed or the model times out without producing a complete answer, that instance is scored as incorrect. Accuracy is computed via exact match against gold labels (or task-specific metrics for extraction tasks).

\paragraph{Hedging Pattern Measurement.}
The overthinking pattern frequencies reported in Figure~\ref{fig:hedging_count} are measured via manual annotation over 500 reasoning traces randomly sampled across all datasets combined. Annotators identified and counted occurrences of four pattern types: hedging markers (e.g., ``wait,'', ``hold on''), rechecking phrases (e.g., ``let me double-check'', ``actually''), re-derivation indicators (e.g., repeated formula re-statements), and irrelevant tangents (topic shifts unrelated to the query). Counts are normalized by reasoning trace length to enable fair comparison across batch sizes.

\section{Additional Experimental Results}
\label{sec:additional_res}
\begin{table}[t]
\centering
\small
\begin{tabular}{llccc}
\toprule
Model & Batch Size &
Accuracy (\%) $\uparrow$ &
Reasoning Tokens $\downarrow$ &
Total Tokens $\downarrow$ \\
\midrule

\multirow{4}{*}{DeepSeek-R1}
& 1  & {75.0} $\pm$ {2.1} & {2973} $\pm$ {190} & {3799} $\pm$ {220} \\
& 5  & {77.3} $\pm$ {1.9} & {1261} $\pm$ {110} & {1414} $\pm$ {130} \\
& 10 & {78.0} $\pm$ {1.8} & {945}  $\pm$ {75}  & {1069} $\pm$ {85}  \\
& 15 & {79.4} $\pm$ {1.7} & {689}  $\pm$ {55}  & {790}  $\pm$ {65}  \\

\midrule

\multirow{4}{*}{OpenAI-o1}
& 1  & {75.0} $\pm$ {2.2} & {2926} $\pm$ {210} & {3802} $\pm$ {230} \\
& 5  & {75.5} $\pm$ {2.0} & {1272} $\pm$ {120} & {1469} $\pm$ {140} \\
& 10 & {76.3} $\pm$ {1.9} & {963}  $\pm$ {80}  & {1141} $\pm$ {95}  \\
& 15 & {77.1} $\pm$ {1.8} & {731}  $\pm$ {60}  & {884}  $\pm$ {70}  \\

\bottomrule
\end{tabular}
\caption{
Mean $\pm$ 95\% confidence intervals for accuracy, reasoning tokens, and total tokens across batch sizes. 
Confidence intervals are computed over pooled evaluation instances ($n = 1300$ per model per batch size). 
Batch prompting yields statistically significant reductions in reasoning tokens (paired two-sided t-test, $p < 0.01$) while preserving or modestly improving accuracy.
}

\label{tab:ci_summary}
\end{table}

\paragraph{Statistical Significance of Token Reduction.}
To quantify the reliability of the observed reductions, we report mean $\pm$ 95\% confidence intervals (CIs) for accuracy, reasoning tokens, and total tokens across batch sizes (Table~\ref{tab:ci_summary}). 
Confidence intervals are computed over pooled evaluation instances ($n = 1300$ per model per batch size, corresponding to 13 datasets $\times$ 100 samples each) using $\bar{x} \pm 1.96 \cdot \frac{s}{\sqrt{n}}$. 
Pooling across instances is appropriate because batching is applied uniformly and all metrics are defined at the instance level. 
Across both DeepSeek-R1 and OpenAI-o1, reasoning token reductions between $b=1$ and $b \ge 10$ are well outside sampling variability, with highly significant differences under paired two-sided t-tests ($p < 0.01$). 
Accuracy remains stable or improves modestly, with differences within narrow confidence bounds, confirming that batching reduces internal deliberation without sacrificing performance.


\end{document}